# Algorithm Portfolio Design: Theory vs. Practice

**Carla P. Gomes**
Rome Laboratory*
Rome Lab, NY 13441
gomes@ai.rl.af.mil

**Bart Selman**
AT&T Bell Laboratories
Murray Hill, NJ 07974
selman@research.att.com

## Abstract

Stochastic algorithms are among the best for solving computationally hard search and reasoning problems. The runtime of such procedures is characterized by a random variable. Different algorithms give rise to different probability distributions. One can take advantage of such differences by combining several algorithms into a portfolio, and running them in parallel or interleaving them on a single processor. We provide a detailed evaluation of the portfolio approach on distributions of hard combinatorial search problems. We show under what conditions the portfolio approach can have a dramatic computational advantage over the best traditional methods.

## 1 Introduction

Randomized algorithms are among the best current algorithms for solving computationally hard problem. Most local search methods for solving combinatorial optimization problems have a stochastic component, both to generate an initial candidate solution, as well as to choose among good local improvements during the search. Complete backtrack-style search methods often also use an element of randomness in their value and variable selection in case of ties. The runtime of these algorithms varies per run on the same problem instance, and therefore can be characterized by a probability distribution. The performance of algorithms can also vary dramatically among different problem instances. In this case, we want to consider the performance profile of the algorithm over a spectrum of problem instances.

*Carla P. Gomes works for Rome Laboratory as a Research Associate.

Given the diversity in performance profiles among algorithms, various approaches have been developed to combine different algorithms to take into account the computational resource constraints and to optimize the overall performance. These considerations led to the development of anytime algorithms (Dean and Boddy 1988), decision theoretic metareasoning and related approaches (Horvitz and Zilberstein 1996; Russell and Norvig 1995), and algorithm portfolio design (Huberman *et al.* 1997). Despite the numerous results obtained in these areas, so far they have not been exploited much by the traditional communities that study hard computational problems, such as operations research (OR), constraint satisfaction (CSP), theorem proving, and the experimental algorithms community.

In order to bridge this gap, we study the possibility of combining algorithms in the context of the recent results concerning the inherent complexity of computationally hard search and reasoning problems. We will provide a rigorous empirical study of the performance profiles of several of the state-of-the-art search methods on a distribution of hard search problems. Our search problems are based on the so-called quasigroup completion task, defined below. For this particular combinatorial search problem, we can vary the computational difficulty and the amount of inherent problem structure in a controlled manner. This enables us to study different aspects of the algorithm performance profiles.

Our studies reveal that in many cases the performance of a single algorithm *dominates* all others, on the problem class under consideration. This may be due to the fact that heuristics are often highly tuned for particular problem domains. Having a single algorithm that dominates over the whole spectrum of problem instances prevents any possible payoff of combining different algorithms. However, we also identify several interesting problem classes where no single method dominates. We will show that on those problem classes, designing a portfolio of several algorithms gives a dramatic improvement in terms of overall performance. In addition, we also show that a good strategy for designing a portfolio is to combine many short runs of the *same* algorithm. The effectiveness of such portfolios explains the common practice of "restarts" for stochastic procedures, where the same algorithm is run repeatedly with different initial seeds for the random number generator. (For related work on the effectiveness of restarts, see *e.g.*, Aldous and Vazirani 1994; Ertel 1991; Selman and Kirkpatrick 1996.)

Our results suggest that the various ideas on flexible computation can indeed play a significant role in algorithm design, complementing the more traditional methods for computationally hard search and reasoning problems.

The paper is organized as follows. In the next section, we introduce our benchmark problem domain: the quasigroup completion problem. We also discuss the theoretical complexity of the problem. In section 3, we give the performance distribution profiles for several complete stochastic search methods on our problem domain. Section 4, we design and evaluate various algorithm portfolios. In section 5, we summarize our results and discuss future directions.

## 2 A Structured Hard Search Problem

In order to study the performance profile of different search strategies, we derive generic distributions of hard combinatorial search problems from the domain of finite algerbra. In particular, we consider the quasigroup domain. A quasigroup is an ordered pair $(Q, \cdot)$, where $Q$ is a set and $(\cdot)$ is a binary operation on $Q$ such that the equations $a \cdot x = b$ and $y \cdot a = b$ are uniquely solvable for every pair of elements $a, b$ in $Q$. The *order* $N$ of the quasigroup is the cardinality of the set $Q$. The best way to understand the structure of a quasigroup is to consider the $N$ by $N$ multiplication table as defined by its binary operation. The constraints on a quasigroup are such that its multiplication table defines a *Latin square*. This means that in each row of the table, each element of the set $Q$ occurs exactly once; similarly, in each column, each element occurs exactly once (Denes and Keedwell 1974).

An *incomplete* or *partial latin square* $P$ is a partially filled $N$ by $N$ table such that no symbol occurs twice in a row or a column. The *Quasigroup Completion Problem* is the problem of determining whether the remaining entries of the table can be filled in such a way that we obtain a complete latin square, that is, a full multiplication table of a quasigroup. We view the pre-assigned values of the latin square as a *perturbation* to the original problem of finding an arbitrary latin square. Another way to look at these pre-assigned values is as a set of additional problem constraints to the basic structure of the quasigroup.

There is a natural formulation of the problem as a Constraint Satisfaction Problem. We have a variable for each of the $N^2$ entries in the multiplication table of the quasigroup, and we use constraints to capture the requirement of having no repeated values in any row or column. All variables have the same domain, namely the set of elements $Q$ of the quasigroup. Pre-assigned values are captured by fixing the value of some of the variables.

Colbourn (1983) showed the quasigroup completion problem to be NP-complete. In previous work, we identified a clear phase transition phenomenon for the quasigroup completion problem (Gomes and Selman 1997). See Figures 1 and 2. From the figures, we observe that the costs peak roughly around the same ratio (approximately 42% pre-assignment) for different values of $N$. (Each data point is generated using 1,000 problem instances. The pre-assigned values were randomly generated.) This phase transition with the corresponding cost profile allows us to tune the difficulty of our problem class by varying the percentage of pre-assigned values.

An interesting application area of latin squares is the design of statistical experiments. The purpose of latin squares is to eliminate the effect of certain systematic dependency among the data (Denes and Keedwell 1974). Another interesting application is in scheduling and timetabling. For example, latin squares are useful in determining intricate schedules involving pairwise meetings among the members of a group (Anderson 1985). The natural perturbation of this problem is the problem of completing a schedule given a set pre-assigned meetings.

The quasigroup domain has also been extensively used in the area of automated theorem proving. In this community, the main interest in this domain has been driven by questions regarding the existence and nonexistence of quasigroups with additional mathematical properties (Fujita et al. 1993; Lam et al. 1989).

## 3 Computational Cost Profiles

We will now consider the computational cost of solving the completion problem for different search strategies. As our basic search procedure, we use a complete backtrack-style search method. The performance of such procedures can vary dramatically depending on the way one selects the next variable to branch on (the "variable selection strategy") and in what order the possible values are assigned to a variable (the "value selection strategy"). There is a large body of work in both the CSP and OR communities exploring different search strategies.

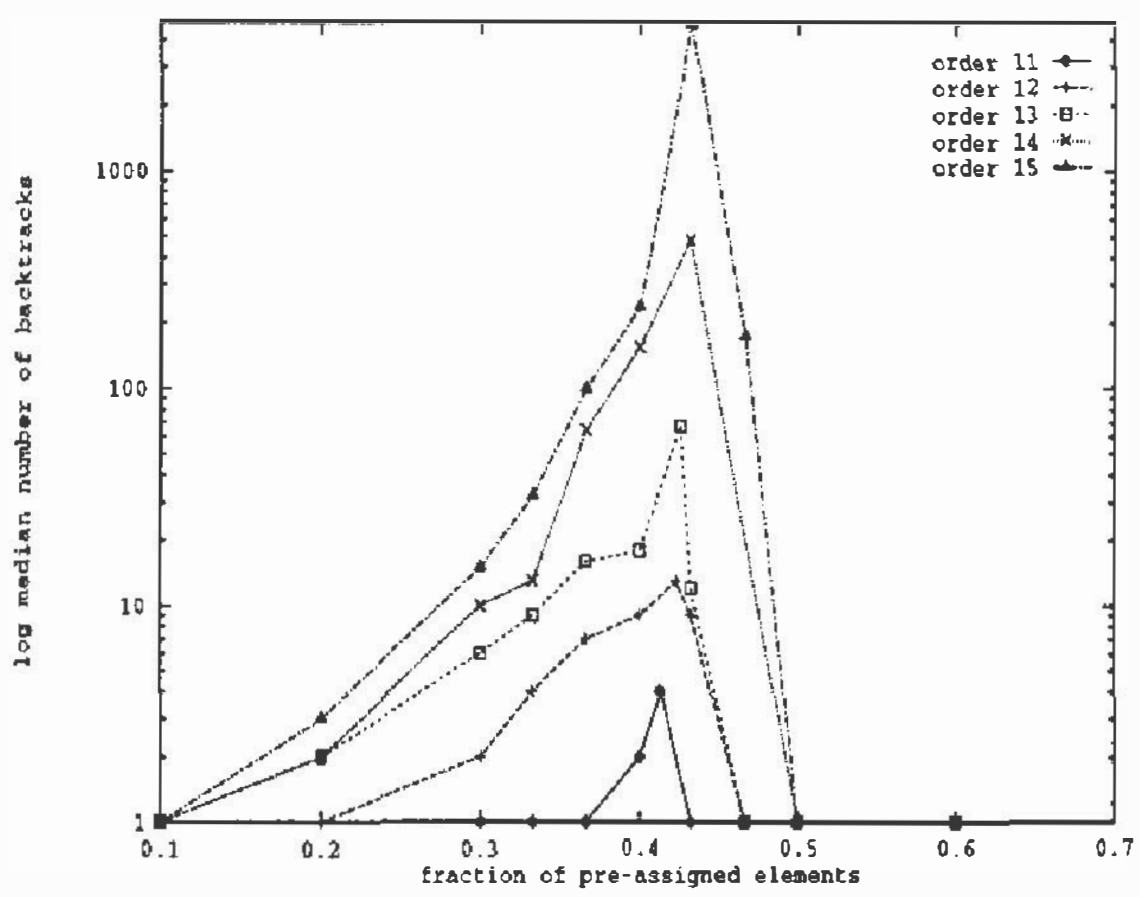

Figure 1: The Complexity of Quasigroup Completion (Log Scale)

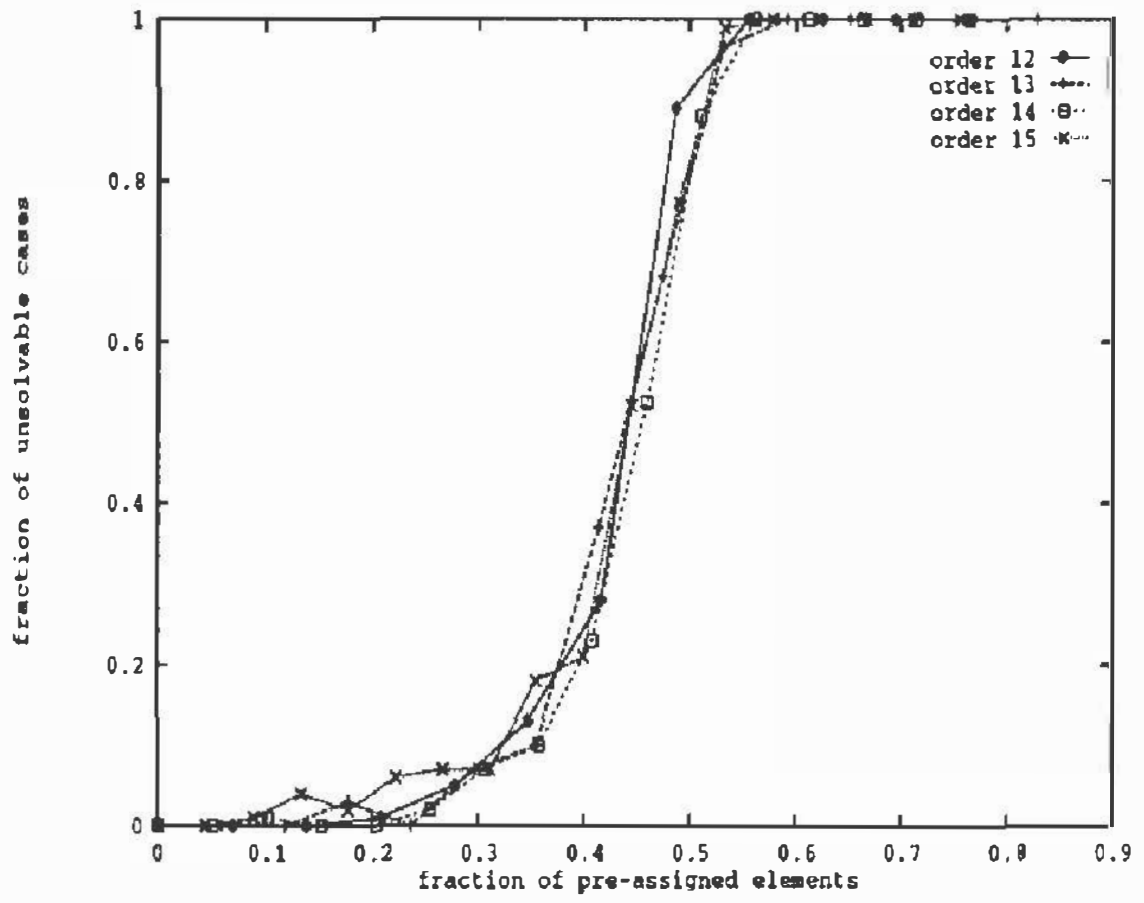

Figure 2: Phase Transition for the Completion Problem

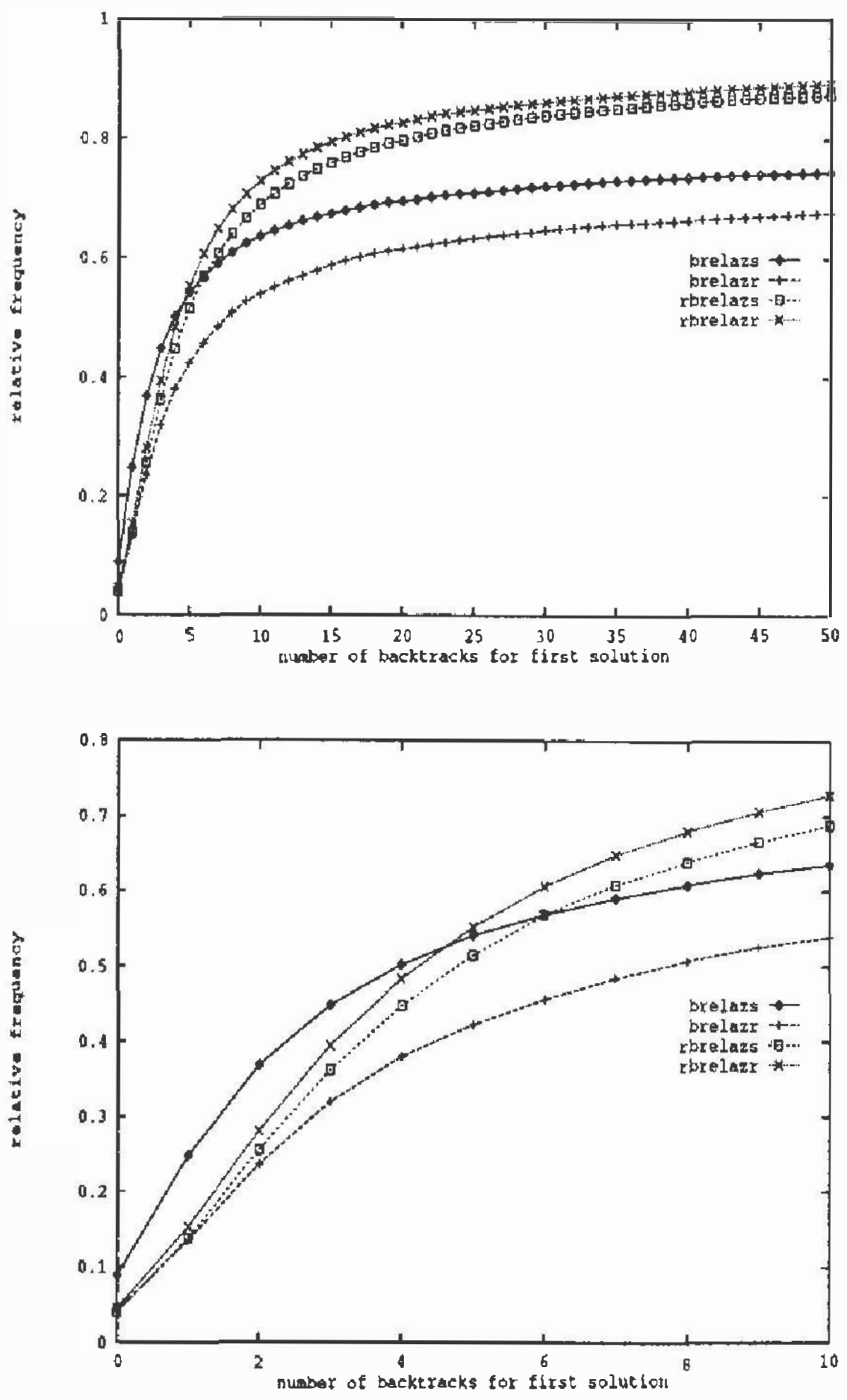

Figure 3: Finding quasigroups of order 20 (no pre-assigned values).

One of the most effective strategies is the so-called First-Fail heuristic.[1] In the First-Fail heuristic, the next variable to branch on is the one with the smallest remaining domain (*i.e.*, in choosing a value for the variable during the backtrack search, the search procedure has the fewest possible options left to explore — leading to the smallest branching factor). We consider a popular extension of the First-Fail heuristic, called the Brelaz heuristics (Brelaz 1979). The Brelaz heuristic was originally introduced for graph coloring procedures. It provides one of the most powerful graph-coloring and general CSP heuristics (Trick and Johnson 1996).

The Brelaz heuristic specifies a way for breaking ties in the First-fail rule: If two variables have equally small remaining domains, the Brelaz heuristic proposes to select the variable that shares constraints with the largest number of the remaining unassigned variables. A natural variation on this tie-breaking rule is what we call the "reverse Berlaz" heuristic, in which preference is given to the variable that shares constraints with the *smallest* number of unassigned variables. Any remaining ties after the (reverse) Brelaz rule are resolved randomly. One final issue left to specify in our search procedure is the order in which the values are assigned to a variable. In the standard Brelaz, value assignment is done in lexicographical order (*i.e.*, systematic). In our experiments, we consider four stragies:

[1]It's really a prerequisit for any reasonable bactrack-style search method. In theorem proving and Boolean satisfiability, the rule corresponds to the powerful unit-propagation heuristic.

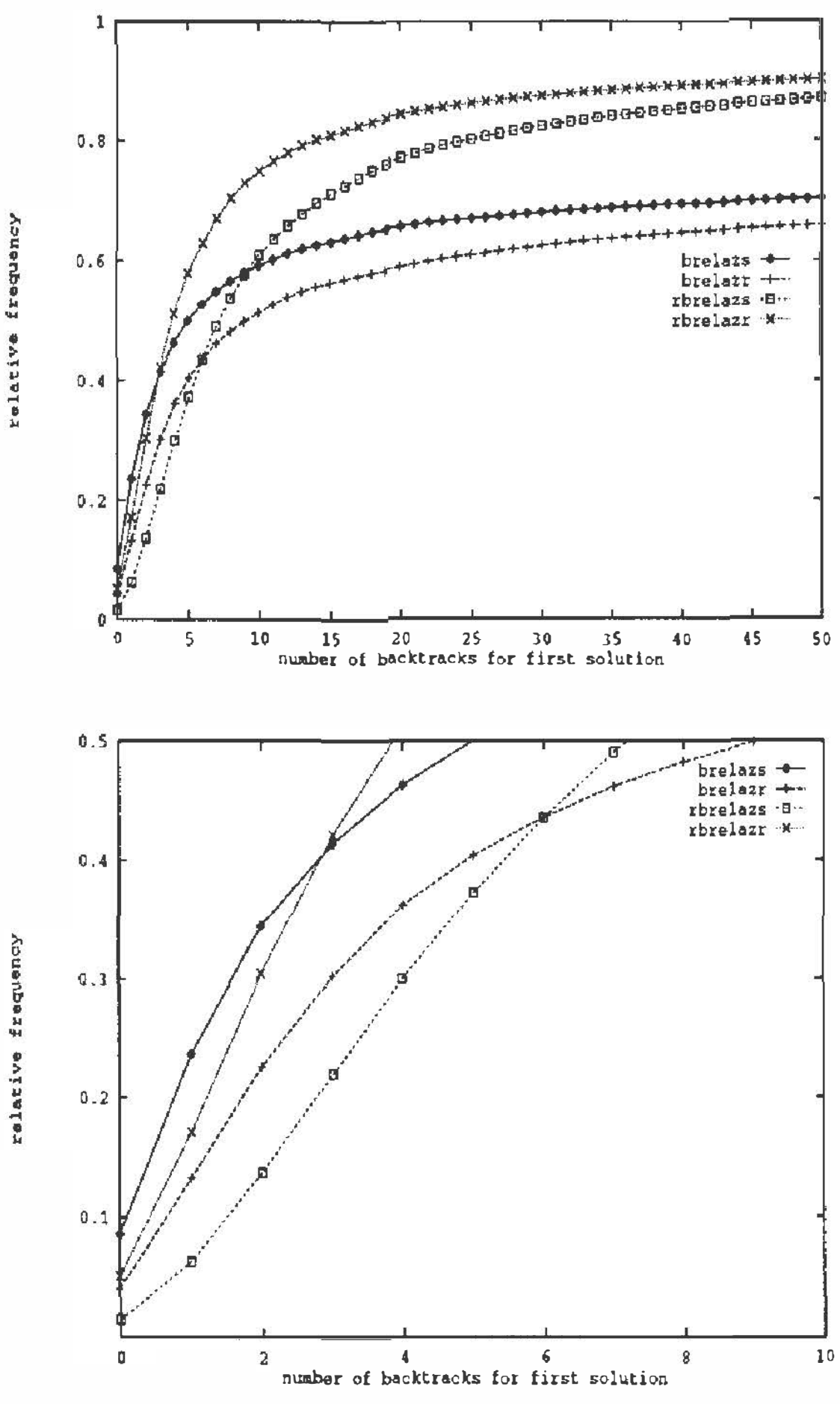

Figure 4: Finding quasigroups of order 20 with 10% pre-assigned values.

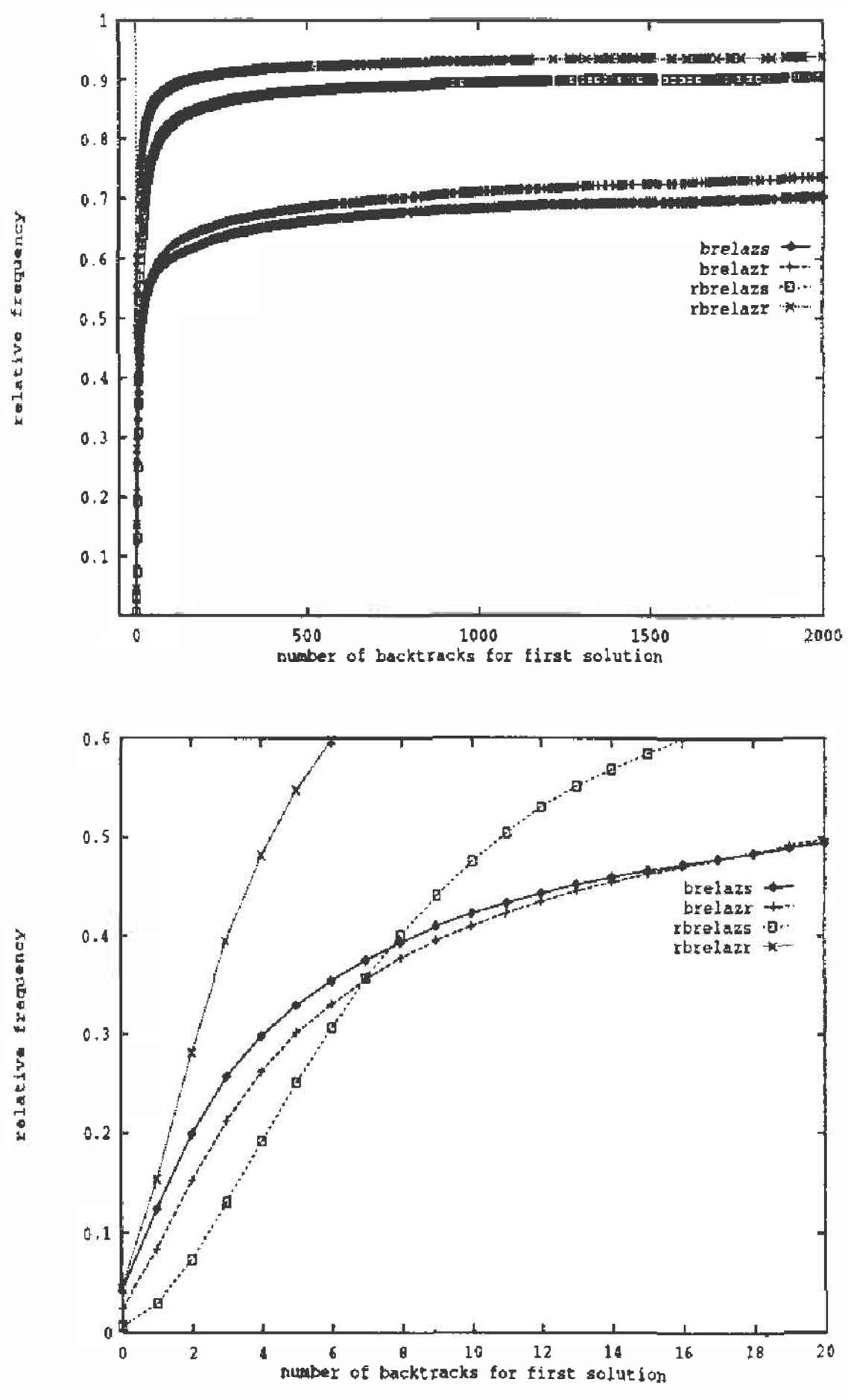

Figure 5: Finding quasigroups of order 20 with 20% pre-assigned values.

- *Berlaz-S* — Berlaz with systematic value selection,
- *Berlaz-R* — Berlaz with random value selection,
- *R-Berlaz-S* — Reverse Berlaz with systematic value selection, and
- *R-berlaz-R* — Reverse Brelaz with random value selection.

Figure 3, shows the performance profile of our four strategies for the problem of finding a quasigroup of order 20 (no pre-assigned values). Each curve gives the cumulative distribution obtained for each strategy by solving the problem 10,000 times. The cost (horizontal axis) is measured in number of backtracks, which is directly proportional to the total runtime of our strategies. For example, the figure shows that R-Berlaz-R, finished roughly 80% of the 10,000 runs in 15 backtracks or less. The top panel of the figure shows the overall profile; the bottom part gives the initial part of the profile.

First, we note that that R-Brelaz-R dominates R-Brelaz-S over the full profile. In other words, the cumulative relative frequency curve for R-Brelaz-R lies above that of R-Brelaz-S at every point along the x-axis. R-Berlaz-S, in turn, strictly dominates Brelaz-R. As we will see below, we often encounter such patterns, where one strategy simply consistently outperforms strategies. Unfortunately, this leaves no room for combining strategies: one simply picks the best strategy. This may explain why some of the ideas about combining algorithms has not received as much attention in the traditional communities that deal with hard computational problems.[2]

From the perspective of combining algorithms, what is most interesting, however, is that in the initial part of the profile (see bottom panel, Figure 3), Brelaz-S dominates the R-Brelaz-R. Intuitively, Brelaz-S is better than R-Berlaz-R at finding solutions *quickly*. However, in the latter part of the cumulative distribution (for more than five backtracks), R-Brelaz-R dominates Brelaz-S. In a sense, R-Brelaz-R gets relatively better when the search gets harder. As we will see in the next section, we can exploit this in our algorithm portfolio design.

[2]There is still the issue of multiple runs with the same method. We'll return to this below.

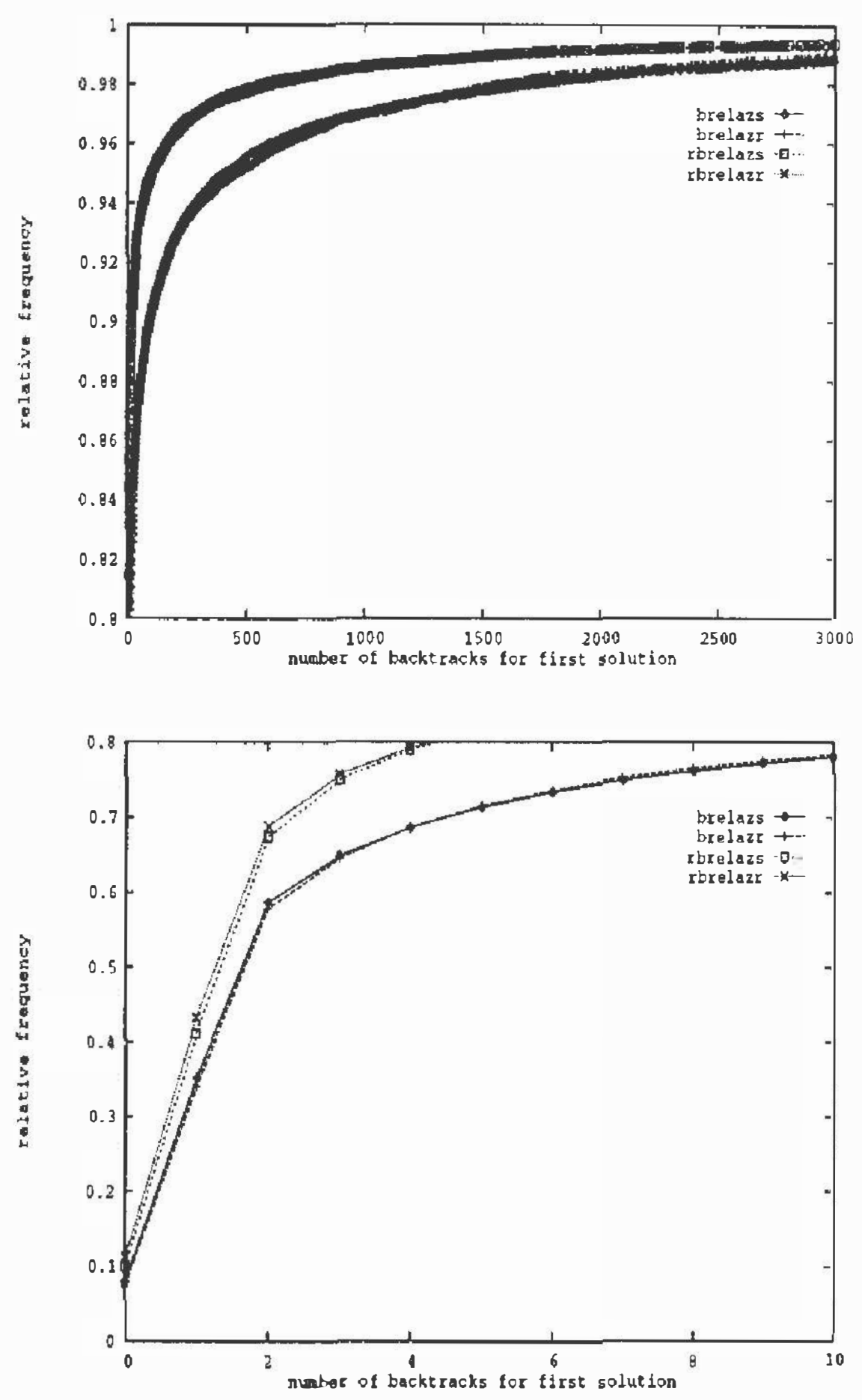

Figure 6: Finding quasigroups of order 10 at the phase transition.

Figure 4, shows the performance profiles for quasigroups with 10% pre-assigned values. We see essentially the same pattern as in Figure 3, but the region where Brelaz-S dominates is relatively smaller. When we increase the percentage of pre-assigned values (20% pre-assigned, Figure 5), we see that R-Brelaz-R completely dominates the other strategies over the whole problem spectrum. This pattern continues for the higher numbers of pre-assigned values (Figure 6, at the phase transition with roughly 40% pre-assigned).

Finally, Figure 7 gives the performance profile for showing the inconsistency of a quasigroup completion problem. The instance in question has 43% preassigned values. Here we again obeserve that Brelaz-S is somewhat better at finding inconsistencies quickly but again R-Brelaz-R dominates for most of the profile. Again, the good initial performance of Brelaz-S can be exploited by combining many short runs, as we will see below.

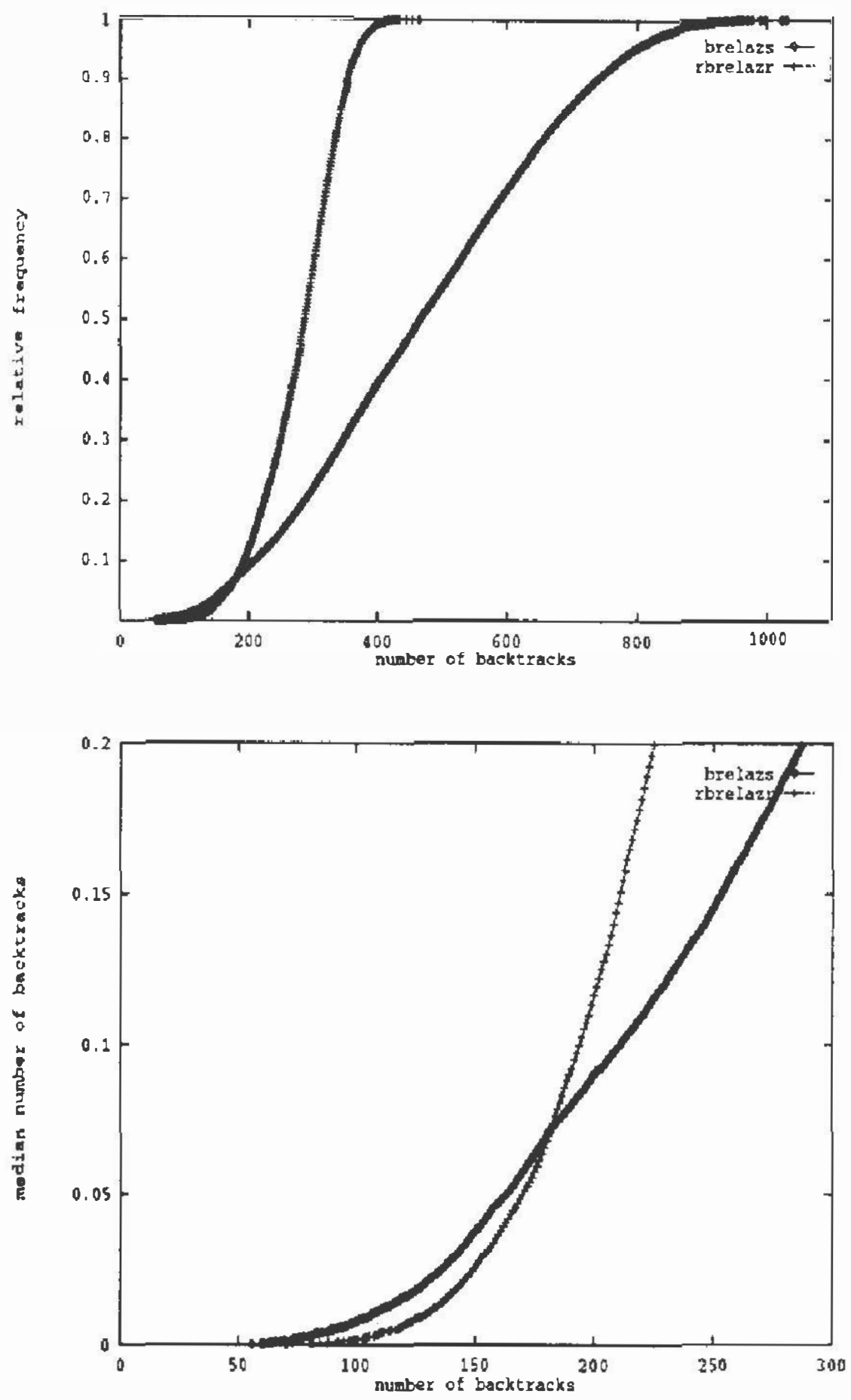

Figure 7: Showing inconsistency of quasigroups completion (order 10 with 43% preassigned values).

## 4 Portfolio Design

A *portfolio of algorithms* is a collection of different algorithms and/or different copies of the same algorithm running on different processors.[3] Here we consider the case of independent runs without interprocess communication.

[3] One can also consider the somewhat more general case of interleaving the execution of algorithms on one or more processors.

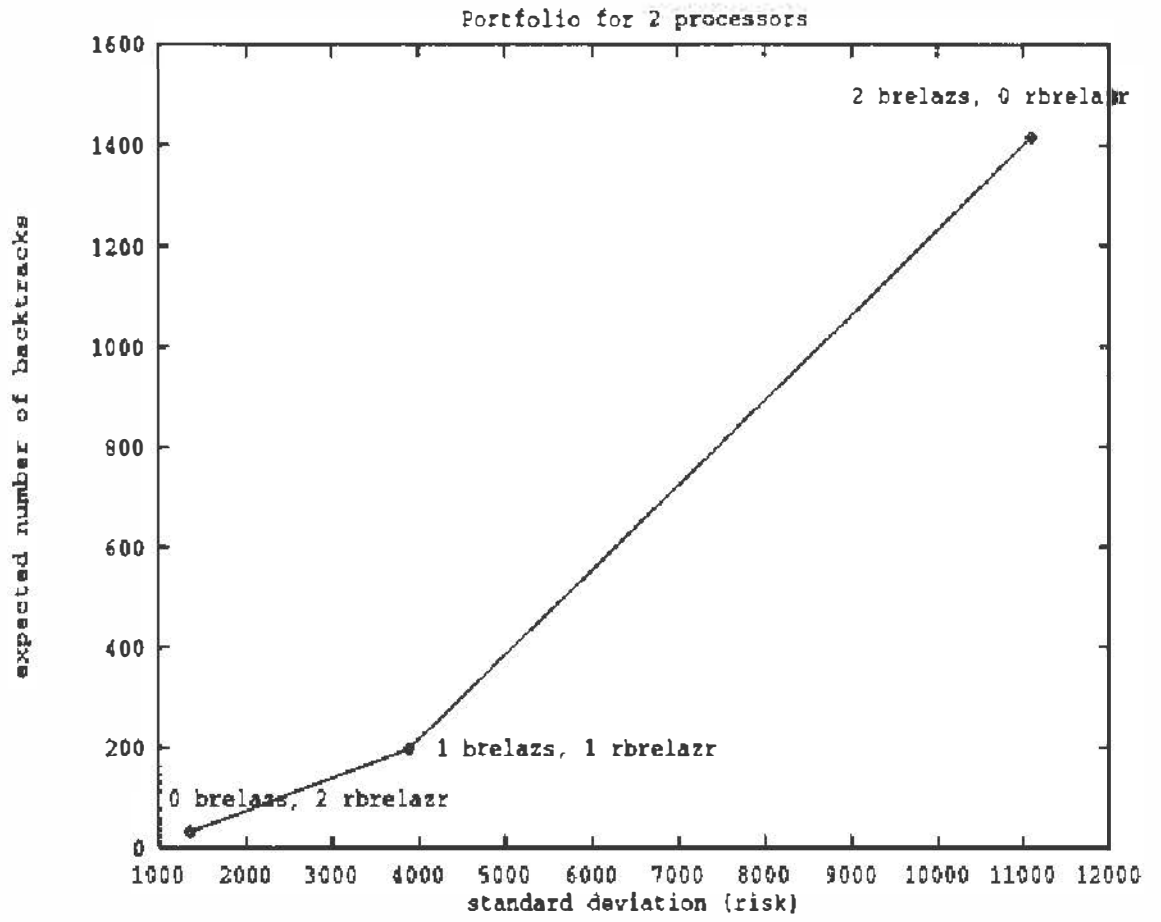

Figure 8: Portfolio for two processors combining Brelaz and R-Brelaz-R.

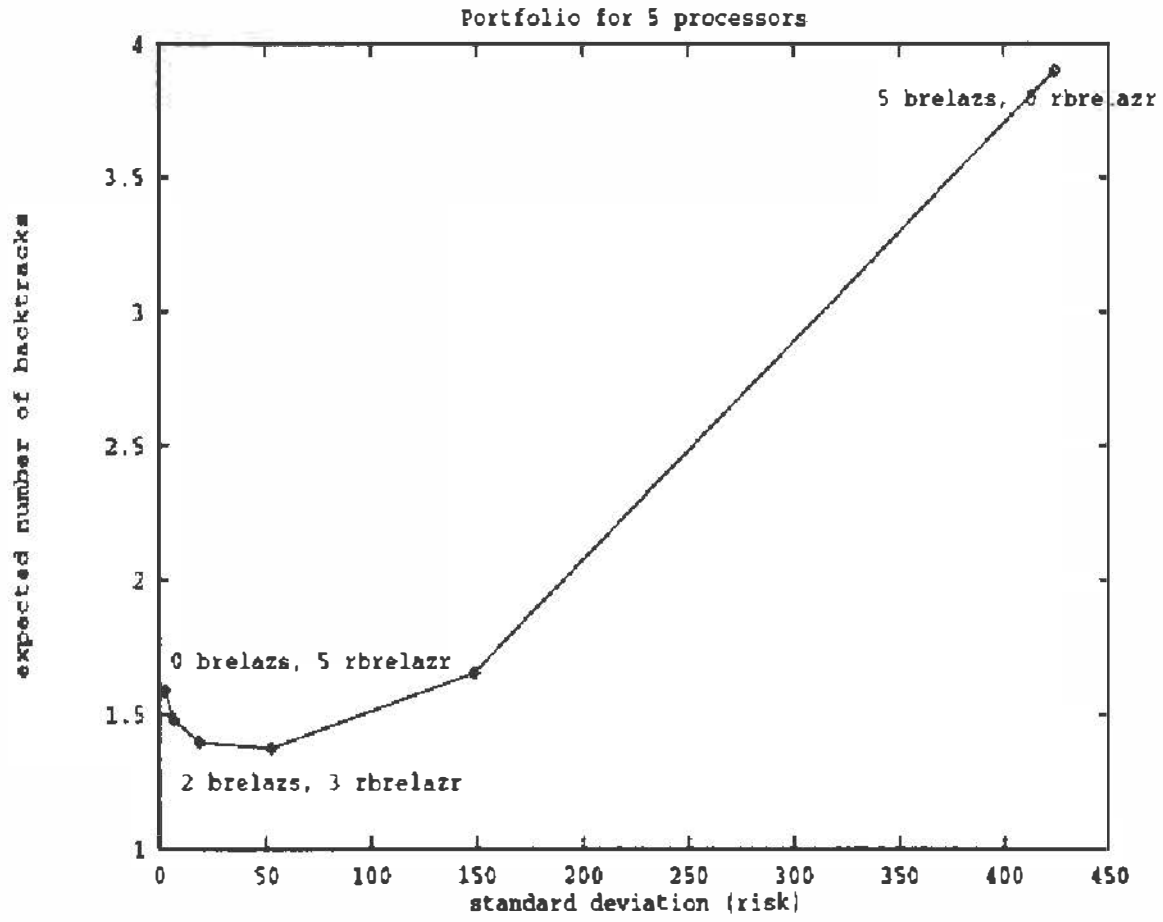

Figure 9: Portfolio for five processors combining Brelaz and R-Brelaz-R.

We are considering Las Vegas type algorithms, *i.e.*, stochastic algorithms that always return a model satisfying the constraints of the search problem or demonstrate that no such model exists (Motwani and Raghavan 1995). The computational cost of the portfolio is therefore a random variable. The *expected* computational cost of the portfolio is simply the expected value of the random variable associated with the portfolio and its *standard deviation* is a measure of the "dispersion" of the computational cost obtained when using the portfolio of algorithms. In this sense, the standard deviation is a measure of the risk inherent to the portfolio.

The main motivation to combine different algorithms into a portfolio is to improve on the performance of the component algorithms, mainly in terms of expected computational cost but also in terms of the overall *risk*. As we will show, some portfolios are strictly preferrable to others, in the sense that they provide a lower risk and also a lower expected computational cost. However, in some cases, we cannot identify any portfolio within a set that is the best, both in terms of expected value and risk. This set of portfolios corresponds to the *efficient set* or *efficient frontier*, following terminology used in the theory of mathematical finance. Within this set, in order to minimize the risk, one has to deteriorate the expected value or, in order to improve the expected value of the portfolio, one has to increase the risk.

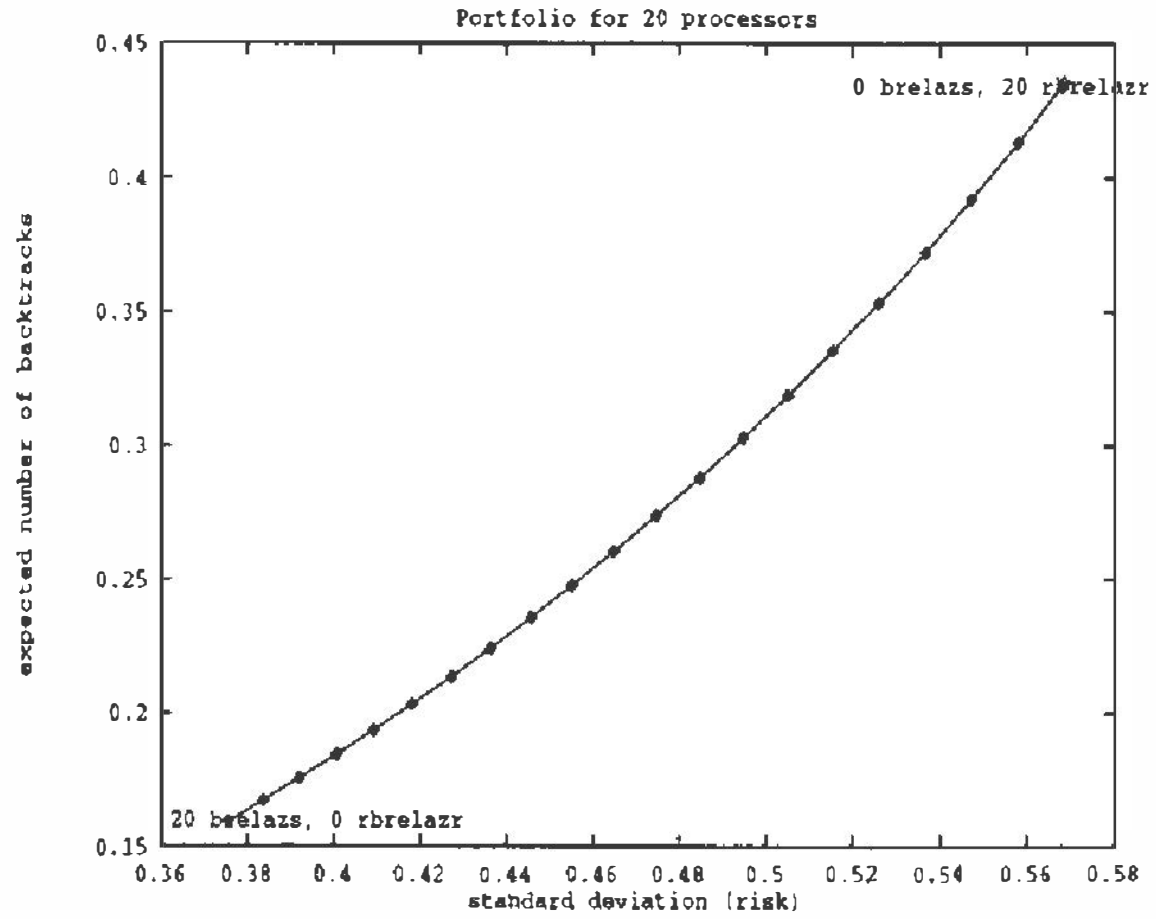

Figure 10: Portfolio for twenty processors combining Brelaz and R-Brelaz-R.

In this context, where we characterize a portfolio in terms of its mean and variance, combining different algorithms into a portfolio only makes sense if they exhibit different probability profiles and none of them *dominates* the others over the whole spectrum of problem instances. As noted earlier, algorithm $A$ dominates algorithm $B$ if the cumulative frequency distribution of algorithm $A$ lies above the cumulative frequency distribution of algorithm $B$ for all points.[4]

Let us consider a set of two algorithms, *algorithm* 1 and *algorithm* 2. Let us associate a random variable with each algorithm: A1 — the number of backtracks that algorithm 1 takes to find the first solution or to prove that a solution does not exist; A2 — the number of backtracks that algorithm 2 takes to find the first solution or to prove that a solution does not exist.

Let us assume that we have $N$ processors and that we design a portfolio using $n1$ processors with *algorithm* 1 and $n2$ processors with *algorithm* 2. So, $N = n1 + n2$. Let us define the random variable associated with this portfolio: X — the number of backtracks that the portfolio takes to find the first solution or to prove that a solution does not exist.

The probability distribution of $X$ is a "weighted" probability distribution of the probability distributions of *algorithm* 1 and *algorithm* 2. More precisely, the probability that $X = x$ is given by the probability that one processor takes exactly $x$ backtracks and all the other ones take $x$ or more backtracks to find a solution or to prove that a solution does not exist.

[4] Another criterion for combining algorithms into a portfolio is given by the algorithm *covariance*.

Let us assume that we have N processors and our portfolio consists of $N$ copies of *algorithm* 1. In this case, P[X=x] is given by the probability that one processor take exactly $x$ backtracks and the other $N-1$ take more than $x$ backtracks, plus the probability that two processors take exactly $x$ backtracks and the other (N-2) one takes more than $x$ backtracks, etc., plus the probability that all the processors take exactly $x$ backtracks to find a solution or to prove that a solution does not exist. The following expression gives the probability function for such a portfolio.

Given $N$ processors, and let $n1 = N$ and $n2 = 0$. P[X=x] is given by

$$\sum_{i=1}^{N} \binom{N}{i} P[A1 = x]^i P[A1 > x]^{(N-i)}$$

To consider two algorithms, we have to generalize the above expression, considering that $X = x$ can occur just within the processors that use *algorithm* 1, or just within the processors that use *algorithm* 2 or within both. As a result, the probability function for a portfolio with two algorithms, is given by the following expression:

Given $N$ processors, $n1$ such that $0 <= n1 <= N$, and $n2 = N - n1$, P[X=x] is given by

$$\sum_{i=1}^{N} \sum_{i'=0}^{n1} \binom{n1}{i'} P[A1 = x]^{i'} P[A1 > x]^{(n1-i')} \times$$

$$\binom{n2}{i''} P[A2 = x]^{i''} P[A2 > x]^{(n2-i'')}]$$

The value of $i''$ is given by $i'' = i - i'$, and the term in the summation is 0 whenever $i'' < 0$ or $i'' > n2$.

In the case of a portfolio involving two algorithms the probability distribution of the portfolio is a summation of a product of two expressions, each one corresponding to one algorithm. In the case of a portfolio comprising M different algorithms, this probability function can be easily generalized, by having a summation of a product of M expressions, each corresponding to an algorithm.

Once we derive the probability distribution for the random variable associated with the portfolio, the calculation of the its *expected value* and *standard deviation* is straightforward.

### 4.1 Empirical results for portfolio design

We now design different portfolios based on our performance profiles from Section 3. We focus on the case of finding a quasigroup of order 20 with no-preassigned values. The performance profiles are given in Figure 3. Note that this is an interesting case from the portfolio design perspective because Brelaz-S dominates in the initial part of the distribution, whereas R-Brelaz-R dominates in the latter part.

Figures 8, 9, and 10 give the expected values and the standard deviations of portfolios for 2, 5, and 20 processors, respectively. (Results derived using the formula given above.) We see that for 2 processors (Figure 8), the portfolio consisting of two copies of the R-Brelaz-R has the best expected value *and* the lowest standard deviation. This portfolio dominates the two other 2-processor portfolios.

When we increase the number of processors, we observe an interesting shift in the optimal portfolio mix. For example, for 5 processors, using 2 Brelaz-S gives a better expected value at only a slight increase in the risk (standard deviation) compared to zero Brelaz-S. In this case, the efficient set comprises three portfolios. One with 5 R-Brelaz-R, one with 1 Brelaz-S and 4 R-Brelaz-R, and one with 2 Brelaz-S and 3 R-Brelaz-R. The situation changes even more dramatically if we go to yet more processors. In particular, with 20 processors (Figure 10), the best portfolio corresponds to using all processors to run the Brelaz-S strategy (the lowest expected value and the lowest standard deviation). The intuitive explanantion for this is that by running many copies of Brelaz-S, we have a good chance that at least one of them will find a solution quickly. This result is consistent with the common use of "random restarts" in stochastic search methods in practical applications. Our portfolio analysis also gives the somewhat counter-intuitive result that, even when given two stochastic algorithms, where neither strictly dominates the other, running multiple copies of a single algorithm is preferrable to a mix of algorithms (Figure 8 and Figure 10).

## 5 Conclusions and Future Work

We have provided concrete empirical results showing the computational advantage of a portfolio approach for dealing with hard combinatorial search and reasoning problems as compared to the best more traditional single algorithm methods. Our analysis also showed what properties of the problem instance distributions lead to the largest payoff for using a portfolio approach in practice. Finally, we saw how the use of random restarts of a good stochastic method is often the optimal strategy. These results suggest that ideas developed in the flexible computation community can play a significant role in practical algorithm design.

**Acknowledgments**

We would like to thank Karen Alguire for developing an exciting tool for experimenting with the quasigroup completion problem. We also would like to thank Nort Fowler for many useful suggestions and discussions, and Neal Glassman for suggesting the domain of combinatorial design as a potential benchmark domain. The first author is a research associate with Rome Laboratory and is funded by the Air Force Office of Scientific Research, under the New World Vistas Initiative (F30602-97-C-0037 and AFOSR NWV project 2304, LIRL 97RL005N25).

# References

Aldous, D. and Vazirani, U. (1994). *Proc. of the 35th Symp. on the Found. of Comp. Sci.*, IEEE Press (1994) 492–501.

Andersen, L. (1985). Completing Partial Latin Squares. *Mathematisk Fysiske Meddelelser*, 41, 1985, 23–69.

Brelaz, D. (1979). New methods to color the verices of a graph. *Comm. of the ACM* (1979) 251-256.

Cheeseman, Peter and Kanefsky, Bob and Taylor, William M. (1991). Where the Really Hard Problems Are. *Proceedings IJCAI-91,* 1991, 163–169.

Colbourn, C. (1983). Embedding Partial Steiner Triple Systems is NP-Complete. *J. Combin. Theory* (A) 35 (1983), 100-105.

Dean, T. and Boddy, M. (1988) An analysis of time-dependent planning. *Proc. AAAI-88*, St. Paul, MI (1988) 49–54.

Dechter, R. (1991) Constraint networks. *Encyclopedia of Artificial Intelligence* John Wiley, New York (1991) 276-285.

Denes, J. and Keedwell, A. (1974) Latin Squares and their Applications. *Akademiai Kiado, Budapest, and English Universities Press*, London, 1974.

Ertel, W. (1991) Performance analysis of competitive or-parallel theorem proving. University of Munchen, Techn. report FKI-162-91, 1992.

Fujita, M., Slaney, J., and Bennett, F. (1993). Automatic Generation of Some Results in Finite Algebra *Proc. IJCAI*, 1993.

Freuder, E. and Mackworth, A. (Eds.). *Constraint-based reasoning.* MIT Press, Cambridge, MA, USA, 1994.

Gent, I. and Walsh, T. (1996) The Satisfiability Constraint Gap. *Artificial Intelligence*, 81, 1996.

Gomes, C.P. and Selman, B. (1997) Solution space of structured problems in the presence of perturbations. Submitted for publication (1997).

Horvitz, E. and Klein, A. (1995) Reasoning, metareasoning, and mathematical truth: studies of theorem proving under limited resources. *Proc. of the Eleventh Conference on Uncertainty in Artificial Intelligence (UAI-95)*, August 1995.

Horvitz, E. and Zilberstein S. (1996) (Eds.) *Proceedings of Flexible Computation*, AAAI Fall Symposium, Cambridge, MA, 1996.

Huberman, B.A., Lukose, R.M., and Hogg, T. (1997). An economics approach to hard computational problems. *Science*, 265, 51–54.

Hogg, T., Huberman, B.A., and Williams, C.P. (Eds.) (1996). Phase Transitions and Complexity. *Artificial Intelligence*, 81 (Spec. Issue; 1996)

Kirkpatrick, S. and Selman, B. (1994) Critical Behavior in the Satisfiability of Random Boolean Expressions. *Science*, 264 (May 1994) 1297–1301.

Lam, C., Thiel, L., and Swiercz, S. (1989) *Can. J. Math.*, Vol. XLI, 6, 1989, 1117-1123.

Mitchell, D., Selman, B., and Levesque, H.J. (1989) Hard and easy distributions of SAT problems. *Proc. AAAI-92*, San Jose, CA (1992) 459–465.

Russell, S and Norvig P. (1995) *Artificial Intelligence a Modern Approach.* Prentice Hall, Englewood Cliffs, NJ. (1995).

Selman, B. and Kirkpatrick, S. (1996) Finite-Size Scaling of the Computational Cost of Systematic Search. *Artificial Intelligence*, Vol. 81, 1996, 273–295.

Smith, B. and Dyer, M. Locating the Phase Transition in Binary Constraint Satisfaction Problems. *Artificial Intelligence*, 81, 1996.

Trick, M. and Johnson, D. (Eds.) (1996) *Proc. DIMACS Challenge on Satisfiability Testing, Graph Coloring, and Cliques.* DIMACS Series on Discr. Math., Am. Math. Soc. Press (1996).

Williams, C.P. and Hogg, T. (1992) Using deep structure to locate hard problems. *Proc. AAAI-92,* San Jose, CA, July 1992, 472–277.

Zhang, W. and Korf, R. A Study of Complexity Transitions on the Asymmetric Travelling Salesman Problem. *Artificial Intelligence*, 81, 1996.